\documentclass[10pt,twocolumn,letterpaper]{article}

\usepackage{cvpr}
\usepackage{times}
\usepackage{epsfig,subfigure}
\usepackage{graphicx}
\usepackage{amsmath}
\usepackage{amssymb}

\usepackage[breaklinks=true,bookmarks=false]{hyperref}

\cvprfinalcopy 

\def\cvprPaperID{****} 

\ifcvprfinal\fi
\begin{document}

\title{Deep Sparse Coding for Invariant Multimodal Halle Berry Neurons}

\author{Edward Kim$^{1,2}$, Darryl Hannan$^{1,2}$, Garrett Kenyon$^2$\\
$^1$Department of Computing Sciences, Villanova University, PA\\
$^2$Los Alamos National Laboratory, Los Alamos, NM\\
{\tt\small edward.kim@villanova.edu,dhannan1@villanova.edu,gkeynon@lanl.gov}
}

\maketitle
\thispagestyle{empty}

\begin{abstract}
Deep feed-forward convolutional neural networks (CNNs) have become ubiquitous in virtually all machine learning and computer vision challenges; however, advancements in CNNs have arguably reached an engineering saturation point where incremental novelty results in minor performance gains.  Although there is evidence that object classification has reached human levels on narrowly defined tasks, for general applications, the biological visual system is far superior to that of any computer.  Research reveals there are numerous missing components in feed-forward deep neural networks that are critical in mammalian vision.   The brain does not work solely in a feed-forward fashion, but rather all of the neurons are in competition with each other; neurons are integrating information in a bottom up and top down fashion and incorporating expectation and feedback in the modeling process.  Furthermore, our visual cortex is working in tandem with our parietal lobe, integrating sensory information from various modalities. 

In our work, we sought to improve upon the standard feed-forward deep learning model by augmenting them with biologically inspired concepts of sparsity, top-down feedback, and lateral inhibition.  We define our model as a sparse coding problem using hierarchical layers.  We solve the sparse coding problem with an additional top-down feedback error driving the dynamics of the neural network.  While building and observing the behavior of our model, we were fascinated that multimodal, invariant neurons naturally emerged that mimicked, ``Halle Berry neurons'' found in the human brain.  These neurons trained in our sparse model learned to respond to high level concepts from multiple modalities, which is not the case with a standard feed-forward autoencoder.  Furthermore, our sparse representation of multimodal signals demonstrates qualitative and quantitative superiority to the standard feed-forward joint embedding in common vision and machine learning tasks.

\end{abstract}

\section{Introduction}
In the past several decades, neuroscientists have been studying the response of the brain to sensory input and theorized that humans have neurons that are colloquially known as grandmother cells.  A grandmother cell is a single neuron that responds to a specific concept or object and activates upon seeing, hearing, or sensibly discriminating that entity, such as the person's grandmother.  In 2005, neuroscientists Quiroga et al. \cite{quiroga2005invariant} conducted a study on eight patients that exhibited pharmacologically intractable epilepsy.  These patients had electrodes implanted in their brains, enabling precise recording from multiple isolated neurons in the medial temporal lobe (MTL).  The MTL has many functions including long-term memory, language recognition, and processing sensory input including auditory and visual signals.   Quiroga et al. sought to answer the question of whether MTL neurons represent concept-level information invariant to metric characteristics of images.  In other words, do MTL neurons fire selectively on individuals, landmarks, or objects?  Their results demonstrated that invariant neurons do exist, suggesting a sparse and explicit neural code.  For example, a woman had a \textit{single} neuron fire when shown a picture of Jennifer Aniston, but not on other pictures of people, places, or things.   Another patient had a different neuron fire when shown a picture of Halle Berry, \textit{as well as} the text string ``Halle Berry'', demonstrating sparse neural codes and invariance of a neuron to specific modalities.

Despite this research, on the computational side, neural networks have gradually moved away from biological thematics.  This has largely been due to engineering breakthroughs in the past several years that have transformed the field of computer vision.  Deep feed-forward convolutional neural networks (CNNs) have become ubiquitous in virtually all vision challenges, including classification, detection, and segmentation.  But further engineering of these networks is reaching a saturation point where incremental novelty in the number of layers, activation function, parameter tuning, gradient function, etc., is only producing incremental accuracy improvements.  Although there is evidence that object classification has reached human levels on certain narrowly defined tasks \cite{russakovsky2015imagenet}, for general applications, the biological visual system is far superior to that of any computer.  Research reveals there are numerous missing components in feed-forward deep neural networks that are critical in mammalian vision.   The brain does not work solely in a feed-forward fashion, but rather all of the neurons are in competition with each other; neurons are integrating information in a bottom up and top down fashion and incorporating expectation and feedback into the inference process.  Furthermore, our visual cortex is working in tandem with our parietal lobe, integrating sensory information from various modalities. 

In the following sections, we describe our motivation and research towards improving the standard feed-forward deep learning model by augmenting them with biologically inspired concepts of sparsity, top-down feedback, and lateral inhibition.  Section \ref{sec:sparsecoding} describes the formulation of the problem as a hierarchical sparse coding problem, Section \ref{sec:lateral} explains how our model incorporates lateral inhibition among neurons, and Section \ref{sec:topdown} defines the dynamics of top-down feedback and demonstrates the effects on a toy problem.  While building and observing the behavior of our model, we were fascinated that multimodal, invariant neurons naturally emerged that mimicked the Halle Berry neurons found in the human brain.  We describe the emergence of these invariant neurons and present experiments and results that demonstrate that our sparse representation of multimodal signals is qualitatively and quantitatively superior to the standard feed-forward joint embedding in common vision and machine learning tasks.

\section{Background}

The current feed-forward deep neural network model has been extremely popular and successful in recent years spanning from the seminal works of LeCun et al. \cite{lecun1995convolutional} to  Krizhevsky et al. \cite{krizhevsky2012imagenet} to He et al. \cite{he2016deep} deep residual neural networks.
Inspiration for this model had biological underpinnings in the work from David Marr's functional model of the visual system \cite{marr1979computational} where levels of computation, e.g. primal sketch to 2.5D to 3D representation, mimicked the cortical areas of the primate visual system.  
Indeed, CNNs possess remarkable similarities with the mammalian visual cortex, as shown by Hubel and Wiesel's receptive field experiments \cite{hubel1962receptive}.  

However it's clear that the dense feed-forward model is not the architecture in the visual system.  Jerry Lettvin \cite{gross2002genealogy} first postulated the existence of grandmother cells, hyper specific neurons that responds to complex and meaningful stimuli.   R. Quiroga et al. \cite{quiroga2005invariant} extended this research further by demonstrating the existence of the Jennifer Aniston neuron and Halle Berry neuron, providing evidence for extreme sparsity when processing sensory input.   Quiroga's study also describes the invariance of neurons to modality, i.e. a neuron would fire not only from the picture, but also from the text.  Further evidence of invariant neurons is provided by Giudice et al. \cite{giudice2009programmed} with what are called ``mirror'' neurons.   Mirror neurons are Hebbian-learned neurons that are optimized to support the association of the perception of actions and the corresponding motor program.  The biological motivation for sparsity is clear, and the computational benefits for sparsity are many.  Glorot et al. \cite{glorot2011deep} describes the advantages of sparsity in a neural network.  These include information disentangling, efficient variable-size representation, and evidence that sparse representations are more linearly separable.  Glorot et al. also notes that sparse models are more neurologically plausible and computationally efficient than their dense counterparts.  Goodfellow et al. \cite{goodfellow2009measuring} found that sparsity and weight decay had a large effect on the invariance of a single-layer network.  As a prime example, Le et al. \cite{le2012building} was able to build distinctive high-level features using sparsity in unsupervised large scale networks.  Lundquist et al. \cite{lundquist2017sparse} was able to use sparse coding to successfully infer depth selective features and detect objects in video.  Our computational network is related to these works; we formulate our model as an unsupervised hierarchical sparse coding problem.

Aside from sparsity, other biological concepts are extremely important to consider.  Both lateral and feedback connections are utilized to perform neural computation \cite{roebuck2014role}.  Lateral inhibition was discovered decades ago between neighboring columns in the visual cortex \cite{blakemore1970lateral,blakemore1972lateral}.  Early visual neurons (for example in V1 - the primary visual cortex) do not act as simple linear feature detectors as they do in artificial neural networks \cite{petro2014contributions}.  They transform retinal signals and integrate top-down and lateral inputs, which convey prediction, memory, attention, expectation, learning, and behavioral context. Such higher processing is fed back to V1 from cortical and subcortical sources \cite{muckli2013network}.  Top-down feedback is also thought to transmit Bayesian inferences of forthcoming inputs into V1 to facilitate perception and reinforce the representation of rewarding stimuli in V1 \cite{kafaligonul2015feedforward}.  In fact, there are many more feedback connections in the brain than there are feed-forward.  There is a clear motivation to include some sort of feedback connection in deep neural networks.  Research in this area is being conducted, including capsule networks providing expectation to lower layers \cite{sabour2017dynamic} and feedback loops in neural circuits \cite{real2017neural}.   For our model, we build in both lateral inhibition and top-down feedback in a simple and elegant way.  

 \begin{figure*}[t]
	\centering
	\subfigure[Deep Sparse Coding model]{\includegraphics[width=8.5cm]{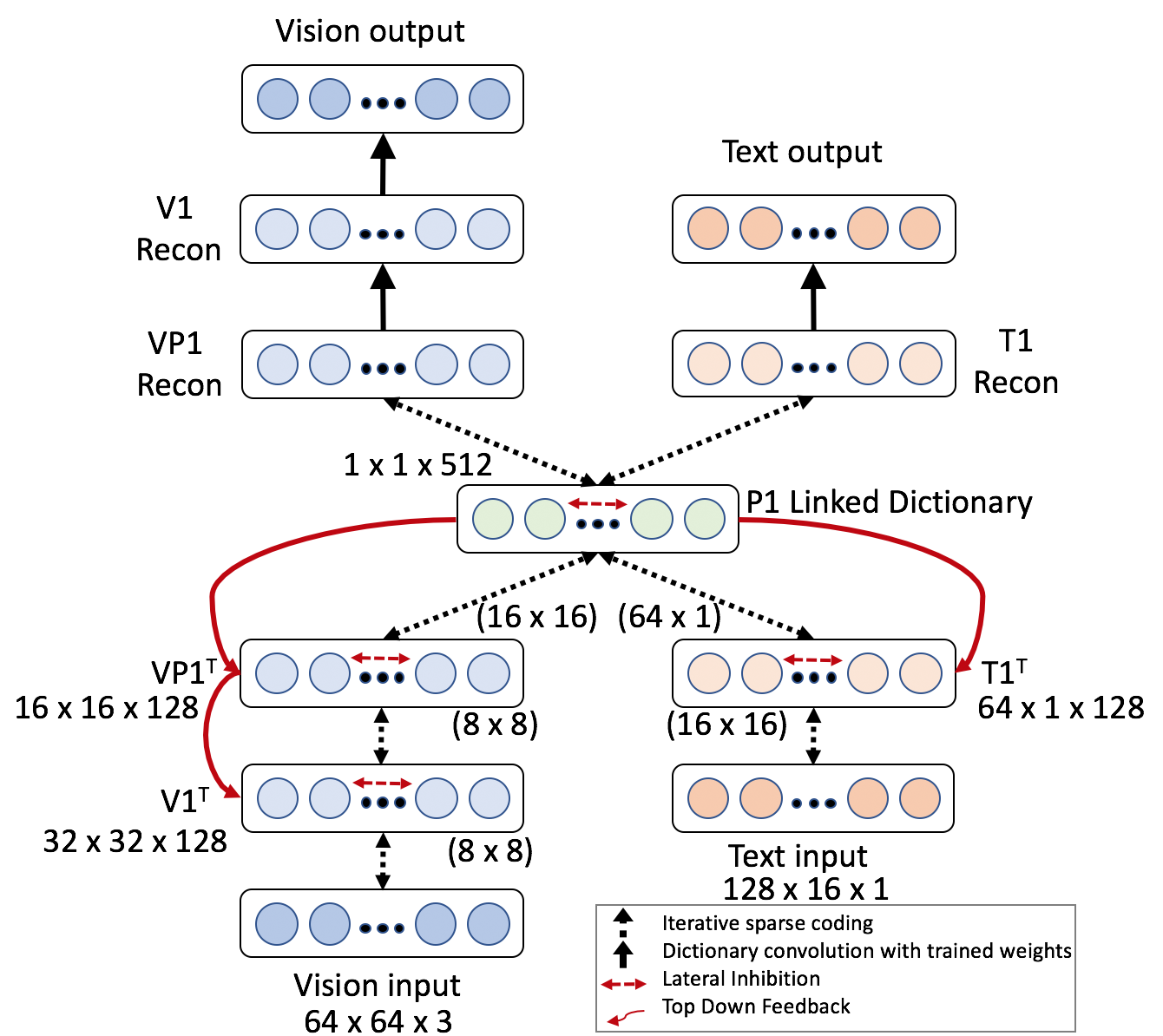}}
	\subfigure[Deep Feed-Forward Autoencoder]{\includegraphics[width=8.4cm]{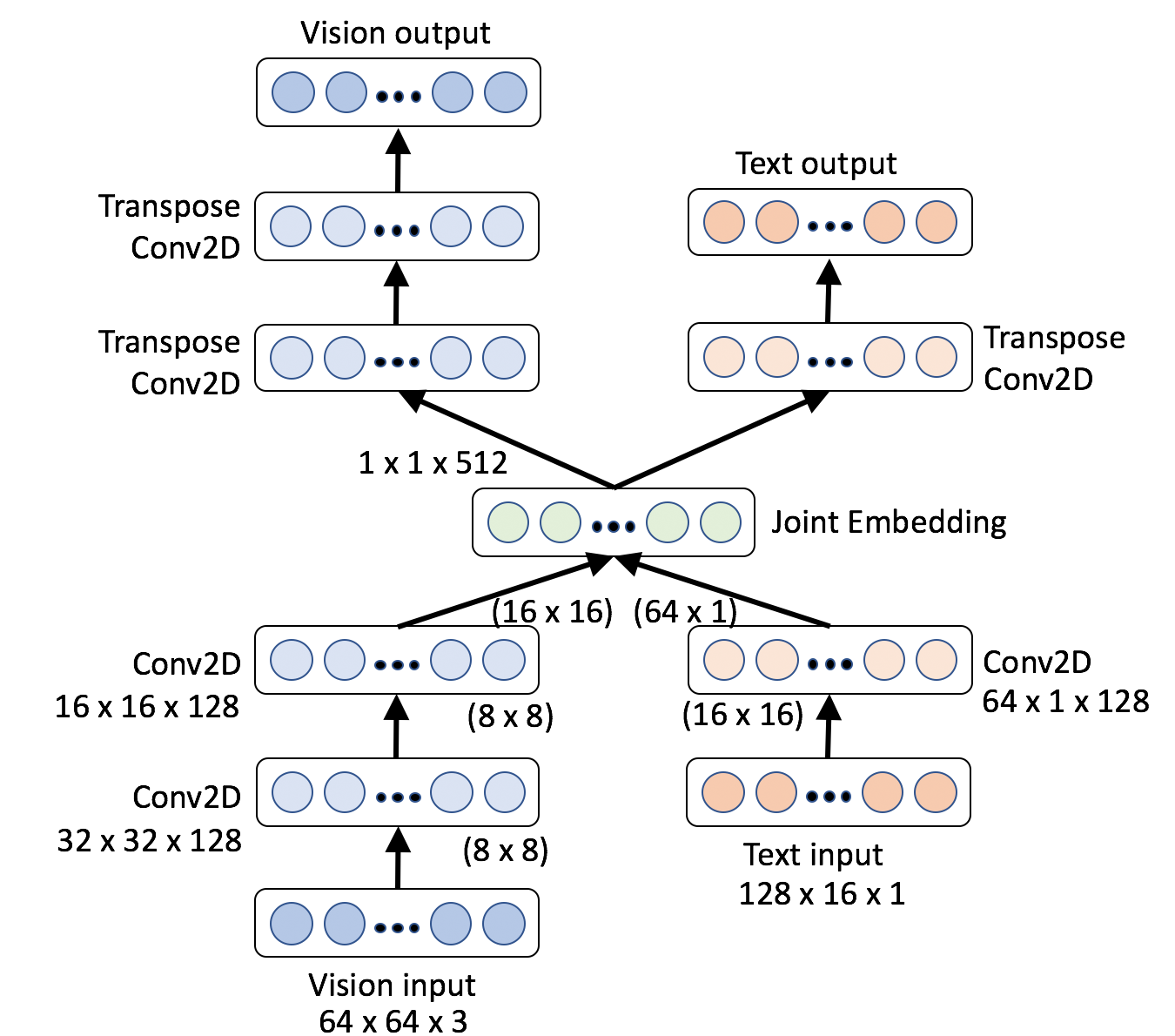}}
\caption{Illustration of the (a) deep sparse coding model and competing (b) feed forward autoencoder model.   In (a), our multimodal representation alternates between the optimization of a sparse signal representation and optimization of the dictionary elements.  Within each layer, the representation is influenced by lateral inhibition from the competing neurons as well as top down feedback from the hierarchical layers.  The reconstruction of the inputs are straightforward convolutions after an internal threshold of the membrane potentials (activations).  Then in (b) we create an equivalent architecture using a feed-forward convolutional neural network with an addition layer to combine the two modalities, ReLU activations, and L1 regularization.}
	\label{fig:process}
\end{figure*}

Finally, our model is multimodal as we process two distinct multimedia streams in the same network.  Others have built and trained multimodal models in neural networks mostly incorporating vision and language \cite{antol2015vqa, kiros2014multimodal, ngiam2011multimodal,srivastava2012multimodal}.  Yang et al. \cite{yang2010image} trained a multimodal model for super resolution using high and low resolution images.  Some have experienced problems in their network since there is no explicit objective for the models to discover correlations across the modalities.  Neurons learn to separate internal layer representations for the different modalities.  Generation of missing modalities, when trained on two modalities, has the additional issue that one would need to integrate out the unobserved visible variables to perform inference \cite{srivastava2012multimodal}.   We demonstrate that the neurons in our network learn \textit{invariant, joint} representations between modalities.  We also show missing modality generation is trivial with our sparse coding model.

\section{Methodology}

\subsection{Deep Sparse Coding}
\label{sec:sparsecoding}
We formulate the problem as a sparse coding problem where we are attempting to reconstruct two modalities.  The first modality, the vision input, is a 64x64 RGB image of a person's face.  The second modality is the text of the person's name.  To better simulate the conditions of the real world, we decided not to represent the text as a one-hot vector typically used in language modeling.  Instead, we represent the text as raw input i.e. the text modality is a 128x16 grayscale image of the printed name of the person.  The full model we developed is a highly recurrent multimodal network seen in Figure \ref{fig:process}(a).  Our sparse coding hierarchy consists of three layers for the visual input and two layers for the text input.  The last layer, i.e. the ``P1 Linked Dictionary'', is a joint layer of both vision and text.  We build a standard feed-forward autoencoder with identical architecture as a comparison model seen in Figure \ref{fig:process}(b).   

Our network can be formulated as a multimodal reconstruction minimization problem which can be defined as follows.  In the sparse coding model, we have some input variable $x^{(n)}$ from which we are attempting to find a latent representation $a^{(n)}$ (we refer to as ``activations'') such that $a^{(n)}$ is sparse, e.g. contains many zeros, and we can reconstruct the original input, $x^{(n)}$ as well as possible. Mathematically, a single layer of sparse coding can be defined as,
\begin{equation}
	\min_\Phi \sum^{N}_{n=1} \min_{a^{(n)}} \frac{1}{2} \| x^{(n)} - \Phi a^{(n)}\|^2_2 + \lambda \|a^{(n)}\|_1
	\label{eq:sparsecode}
\end{equation}
Where $\Phi$ is the dictionary, and $\Phi a^{(n)} = \hat{x}^{(n)}$, or the reconstruction of $x^{(n)}$.  The $\lambda$ term controls the sparsity penalty, balancing the reconstruction versus sparsity term.  $N$ is the total training set, where $n$ is one element of training.  
We reformulate the reconstruction of a signal by substituting the $\Phi \cdot a^{(n)}$ term with $\Phi \circledast a^{(n)}$, where $\circledast$ denotes the transposed convolution (deconvolution) operation, and $\Phi$ now represents a dictionary composed of small kernels that share features across the input signal.   

We use the Locally Competitive Algorithm (LCA) \cite{rozell2007locally} to minimize the mean-squared error (MSE) with sparsity cost function as described in Equation \ref{eq:sparsecode}.  The LCA algorithm is a biologically informed sparse solver that uses principles of \textit{thresholding} and \textit{local competition} between neurons.  The LCA model is governed by dynamics that evolve the neuron's internal state when presented with some input image.  The internal state, i.e. ``membrane potential'', charges up like a leaky integrator and when it exceeds a certain threshold, will activate that neuron.  This activation will then send out inhibitory responses to units within the layer to prevent them from firing.  The input potential to the state is proportional to how well the image matches the neuron's dictionary element, while the inhibitory strength is proportional to the activation and the similarity of the current neuron and competing neuron's convolutional patches, forcing the neurons to be decorrelated.  We will derive the lateral inhibition term in the following section.


The idea of thresholding the internal state of a neuron (membrane) is important when building deep, hierarchical sparse coding models.  Sparse coding a signal that is already a sparse code is difficult and challenging from both a mathematical and logical point of view.  Thus, stacked sparse implementations attempt to densify a sparse code before generating another sparse code.  For example, some do a pooling after a sparse layer \cite{le2012building,gwon2016deep,masci2011stacked,cha2015multimodal} or other operations to densify the sparse layer \cite{he2014unsupervised}.

\begin{figure}[h]
	\centering
	\includegraphics[width=6.5cm]{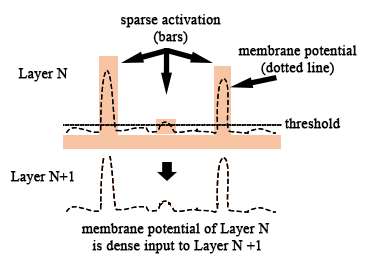}
\caption{Sparse codes can be extracted from any Layer N using a thresholding of the membrane potential.  The membrane potential is passed to the next layer as a dense input for hierarchical sparse coding.}
	\label{fig:dense}
\end{figure}

Our model does not require such operations.  Our reconstruction activation maps are thresholded and sparse; however, the input signal passed to the next hierarchical layer is the dense membrane potential of the current layer.  Figure \ref{fig:dense} illustrates this concept.

\subsection{Lateral Inhibition}
\label{sec:lateral}

The LCA model is an energy based model similar to a Hopfield network \cite{hopfield1984neurons} where the neural dynamics can be represented by a nonlinear ordinary differential equation.  Let us consider a single input image at iteration time $t$, $x(t)$. We define the internal state of the neuron as $u(t)$ and the active coefficients as $a(t) = T_{\lambda}(u(t))$.  The internal state and active coefficients are related by a monotonically increasing function, allowing differential equations of either variable to descend the energy of the network.
The energy of the system can be represented as the mean squared error of reconstruction and a sparsity cost penalty $C(\cdot)$,
\begin{equation}
	E(t) =  \frac{1}{2} \| x(t) - \Phi a(t)\|^2_2 + \lambda \sum_{m}C(a_m(t))
\end{equation} 
Thus the dynamics of each node is determined by the set of coupled ordinary differential equations,
\begin{equation}
\label{eq:origdrive}
 \frac{du_m}{dt} =  -u_m(t) + (\Phi^T x(t))_m - (\Phi^T \Phi a(t) - a(t))_m
\end{equation}
where the equation is related to leaky integrators.  The $-u_m(t)$ term is leaking the internal state of neuron $m$, the $(\Phi^T x(t))$ term is ``charging up'' the state by the inner product (match) between the dictionary elements and input patch, and the $(\Phi^T \Phi a(t) - a(t))$ term represents the inhibition signal from the set of active neurons proportional to the inner product between dictionary elements.  The $- a(t)$ in this case is eliminating self interactions.  In summary, neurons that match the input image charge up faster, then pass a threshold of activation.  Once they pass the threshold, other neurons in that layer are suppressed proportional to how similar the dictionary elements are between competing neurons.  This prevents the same image component from being redundantly represented by multiple nodes.

 \begin{figure*}[t]
	\centering
	\includegraphics[width=17.0cm]{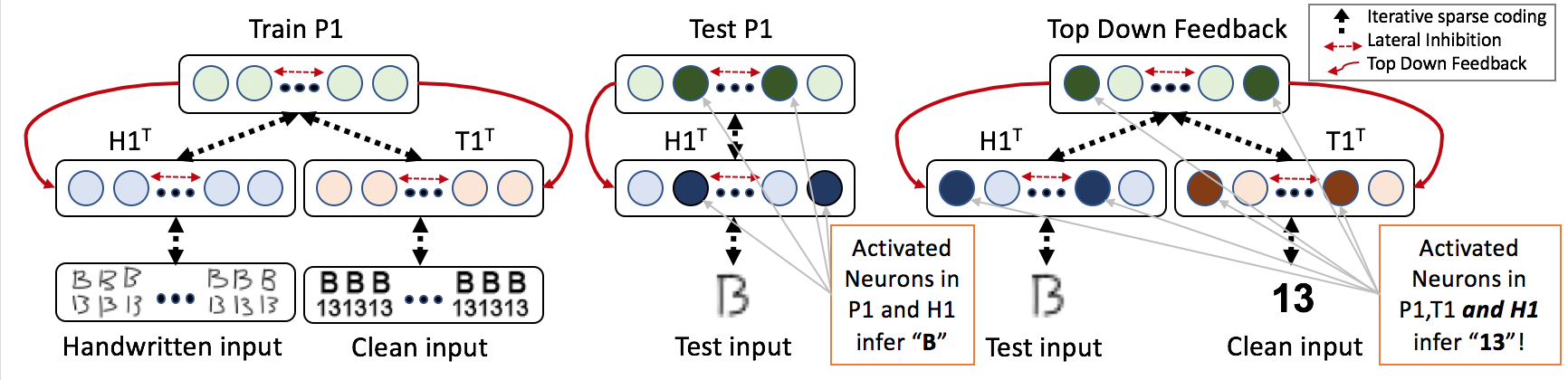}
\caption{An illustrative example of top-down feedback on a toy problem.  We train a small sparse coding network on 50 handwritten characters and 50 printed characters.  The handwritten characters B and 13 can be easily confused as shown in the test input.  Given only the handwritten input, the network speculates whether the character is a B or 13.  Given that the network thinks this is a B, we then provide the network with a contradictory 13 in print.  The network at T1 and P1 naturally change their minds as to the label of the test input, but the most interesting part is that H1 also changes its representation based upon the expectation from top-down feedback.}
	\label{fig:topdowntoy}
\end{figure*}

\subsection{Top-Down Feedback}
\label{sec:topdown}
When we view the world, our brains are continuously trying to understand what it perceives.  Rarely are we confused about our environment, and that is a function of our higher cognitive functions providing feedback to the lower sensory receptors.  If there is some sort of discordance i.e. error between the levels of cognition in our brain, our neural pathways will rectify feed-forward perception, and feed-back error, gradually settling into confident inference.  Similarly, we incorporate top down feedback into the model as the error caused by the upper layers of the hierarchy.  The error propagates down to the lower layers forcing the lower layers to adjust their representation so that it matches both the reconstruction of the signal while mitigating the error of the upper levels.   

In the neural network, the layer above is attempting to reconstruct the given layer's internal state, i.e. membrane potential at layer $N$, $u^{N}(t)$, using the same MSE reconstruction energy.  When adding this term to the given layer, we can take the gradient of the energy with respect to the membrane potential $u^N(t)$ such that,  
\begin{equation}
\label{eq:drive}
	\frac{dE}{du^N(t)} = r^{N+1}(t) = u^N(t) - \Phi^{N+1} a^{N+1}(t)
\end{equation}  
This error can thus be used as an inhibitory signal to the driving mechanics of the lower layer, such that equation \ref{eq:origdrive} is modified to,
\begin{equation}
 \frac{du_m}{dt} =  -u_m(t) + (\Phi^T x(t))_m - (\Phi^T \Phi a(t) - a(t))_m - r^{N+1}(t)
\end{equation}

\begin{figure}[th]
	\centering
	\subfigure[t-SNE of P1]{\includegraphics[height=3.25cm]{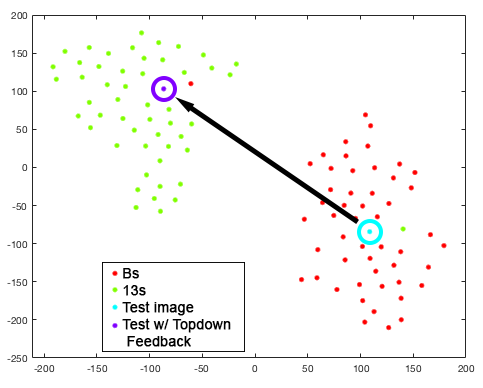}}
	\subfigure[t-SNE of H1]{\includegraphics[height=3.25cm]{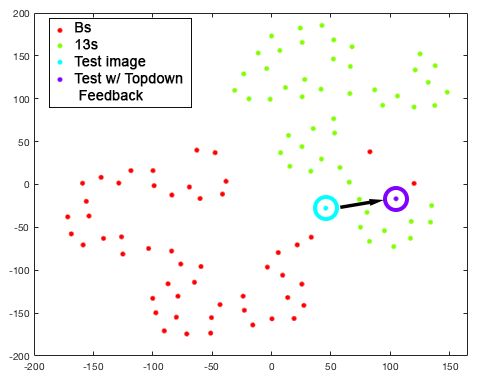}}
\caption{Low dimensional plots of the activation features created by B's and 13's in (a) P1 and (b) H1.  Originally, the test image lies in the middle of the B cluster in P1 and closer to the B cluster in H1.  With the introduction of the printed 13, the test image changes class drastically in P1 to the 13 cluster, and also shifts classes in H1 with top-down feedback.}
	\label{fig:tsnetopdown}
\end{figure}

To further elucidate the concept of top-down feedback in our network, we present a toy example as shown in Figure \ref{fig:topdowntoy}.  Our example is a small sparse coding network that takes in a handwritten modality and its associated printed text.  The network is trained on two types of characters, ``B'' and ``13''.  We train a small sparse coding network on 50 handwritten characters and 50 printed characters.  The handwritten characters B and 13 can be easily confused as shown in the test input.
Based upon nearest neighbor clustering, the test image would be classified as a ``B'' in both P1 and H1 feature space.
The test input image's euclidean distance in P1's 128-dim feature space to class cluster centers is (\textbf{B} = 0.3509 $<$ \textbf{13} = 0.4211).  At H1's 4096-dim feature space, the test image is very much at the border with (\textbf{B} = 0.6123 $<$ \textbf{13} = 0.6131).  Given that the network believes that the input image is a ``B'', we introduce a contrary ``13'' input to the text modality, T1.  The text branch informs P1 of its extremely strong belief that the input image is a ``13'' which drastically sways P1's opinion.  This is no surprise, nor is it a feat of our sparse coding top-down feedback.  Any feed-forward neural network has the ability to change its prediction at a higher level based upon additional information.  What makes our model extraordinary, is that it not only changes its prediction at the high levels, but then forces H1 to conform to its belief.  After top-down feedback, P1 (\textbf{B} = 0.5013 $>$ \textbf{13} = 0.3756) and H1 (\textbf{B} = 0.4271 $>$ \textbf{13} = 0.4020).  This has biological basis \cite{petro2014contributions} where feedback at higher areas in the brain have been shown to suppress predictable inputs in the visual cortex \cite{alink2010stimulus}.  
  We can visualize the effect of top-down feedback by plotting the feature representations at P1 and H1.  We use the t-Distributed Stochastic Neighbor Embedding (t-SNE) \cite{maaten2008visualizing} technique to reduce the dimensionality of the representations to a 2D plot, see Figure \ref{fig:tsnetopdown}.



\section{Experiments and Results}
For our experimentation, we explored the use of faces and names as the two modalities in the Deep Sparse Coding (DSC) model.  We evaluate our model against a standard Feed-Forward Autoencoder (FFA) and perform an analysis in regard to multimodal representation, feature extraction, and missing data generation.  
\subsection{Dataset}
 \begin{figure}[th]
	\centering
	\includegraphics[width=8cm]{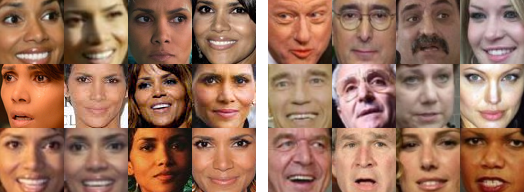}
\caption{Sample images from the Labeled Faces in the Wild dataset.  We augment the dataset with more images of Halle Berry.}
	\label{fig:ss}
\end{figure}

We use a subset of the Labeled Faces in the Wild (LFW) dataset  \cite{Huang2012a}.  This dataset consists of face photographs collected from the web and labeled with the name of the person pictured.  The only constraint was that these faces were detected by the Viola-Jones face detector.  We further augment the faces in the wild dataset with more images of Halle Berry simulating that fact that the more frequent the image, the more likely that it will produce a grandmother cell and an invariant neuron in the model.  The number of images of Halle Berry has been augmented to 370 using Google images and images scraped from IMDB, see Figure \ref{fig:ss}.  However, even with this data augmentation Halle Berry is only the second most frequently represented person; George W Bush remains the most frequent face in the dataset with a total of 530 images.  The total number of faces used for training and testing was 5,763 consisting of 2,189 unique persons.

Each face image is a 64x64x3 color image.  We use the name of the person to create the corresponding image containing the text of the name in 20 point Arial font.  Each text image is a 128x16 grayscale image.

\subsection{Model Implementation}
The Deep Sparse Coding model was implemented using OpenPV\footnote{https://github.com/PetaVision/OpenPV}.  OpenPV is an open source, object oriented neural simulation toolbox optimized for high-performance multi-core computer architectures.  The model was run on an Intel Core i7 with 12 CPUs and an Nvidia GeForce GTX 1080 GPU.  We run the LCA solver for 400 iterations per input.  If we consider one epoch of the training data to be 4000 images, the training time per epoch is approximately 7 hours, or about 6.3 seconds per input image/text pair.
The Feed-forward Autoencoder was implemented using Keras\footnote{https://keras.io/} on top of TensorFlow.  We use convolutional layers with ReLU activation functions and an addition layer to merge the two modalities into one embedding.  Activity regularization using the L1 norm is added to the model.   The time required to run through an epoch of the training data is 47 seconds, or about 0.012 seconds per input image/text pair.  We train the Deep Sparse Coding model for 3 epochs and the Feed Forward Autoencoder for 25 epochs.

 \begin{figure}[th]
	\centering
	\subfigure[\textbf{DSC} Activation Halle Berry.  Note the strong spike on Neuron 326.]{\includegraphics[height=3.75cm]{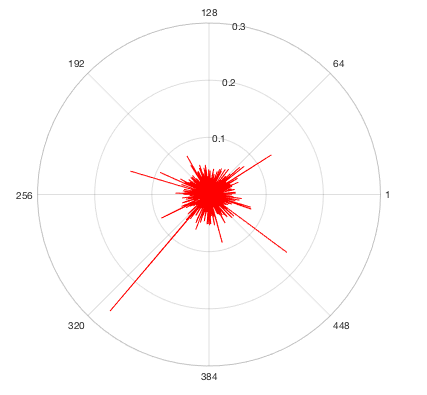}}\hfill
	\subfigure[\textbf{DSC} Activation on Random.  Random input distirbutes activation across neurons.]{\includegraphics[height=3.75cm]{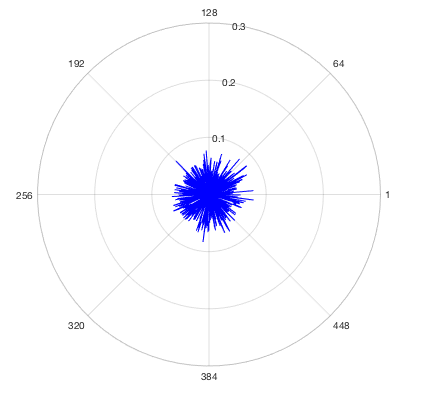}}
	\subfigure[\textbf{FFA} Activation Halle Berry.  Strong spikes on numerous neurons. ]{\includegraphics[height=3.85cm]{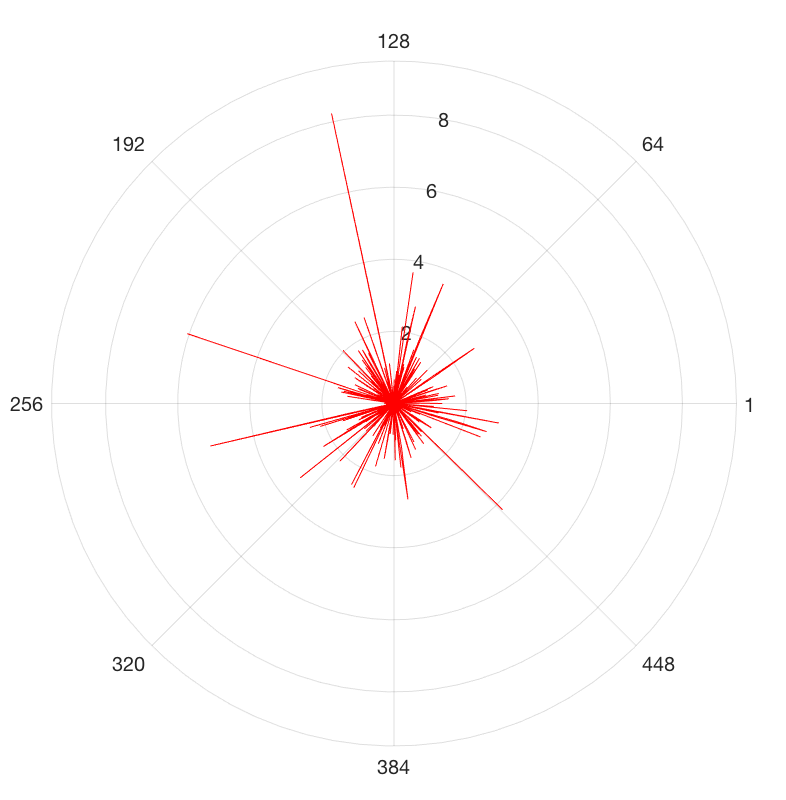}}\hfill
	\subfigure[\textbf{FFA} Activation Random.  Strong spikes on the same neurons as Halle Berry faces.]{\includegraphics[height=3.85cm]{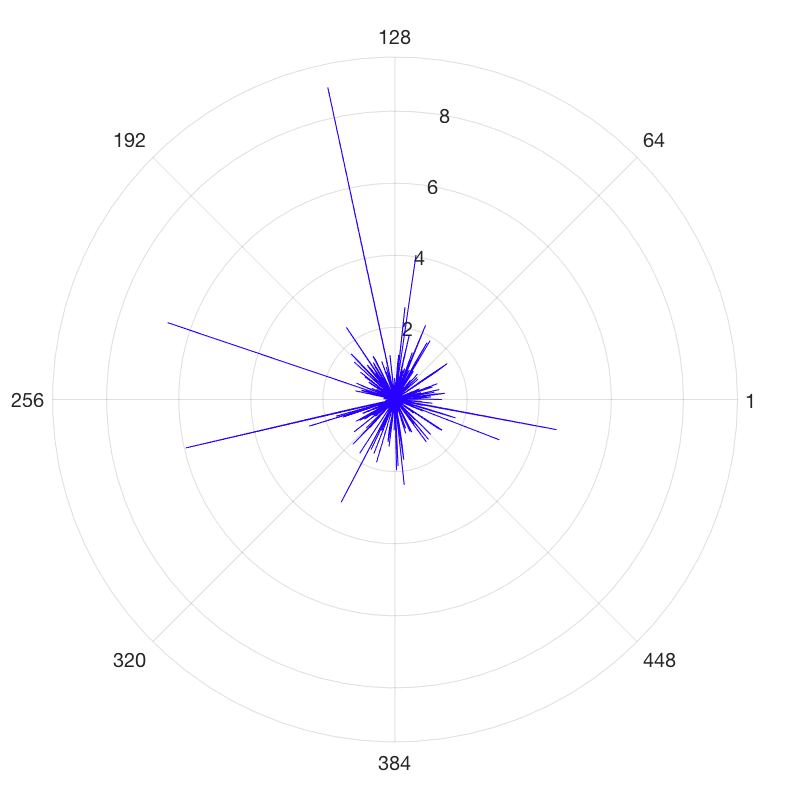}}
\caption{Average activation of faces of 75 Halle Berry and 75 Random faces on the Deep Sparse Coding (DSC) model and the Feed-Forward Autoencoder (FFA) model.  The distinct spike in the DSC model (neuron 326) fires strongly on both the picture and text of Halle Berry; whereas, in the FFA model, the activations are shared by different inputs.  }
	\label{fig:halleberryneuron}
	
\end{figure}
	
 \begin{figure}[th]
	\centering
	\subfigure[N-326 activation on Vision only]{\includegraphics[width=4.0cm]{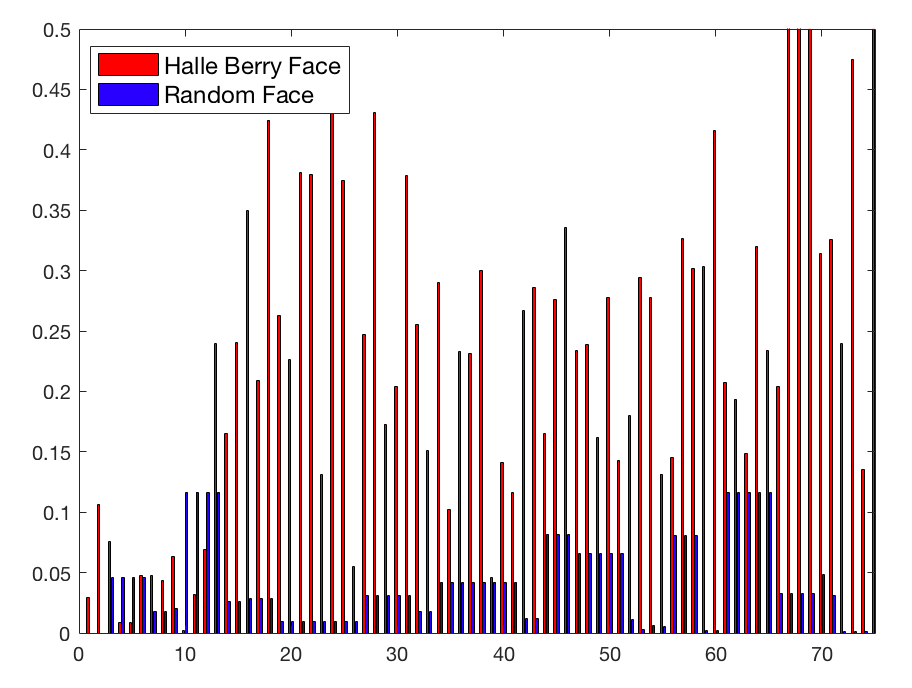}}\hfill
	\subfigure[N-326 activation on Text only]{\includegraphics[width=4.0cm]{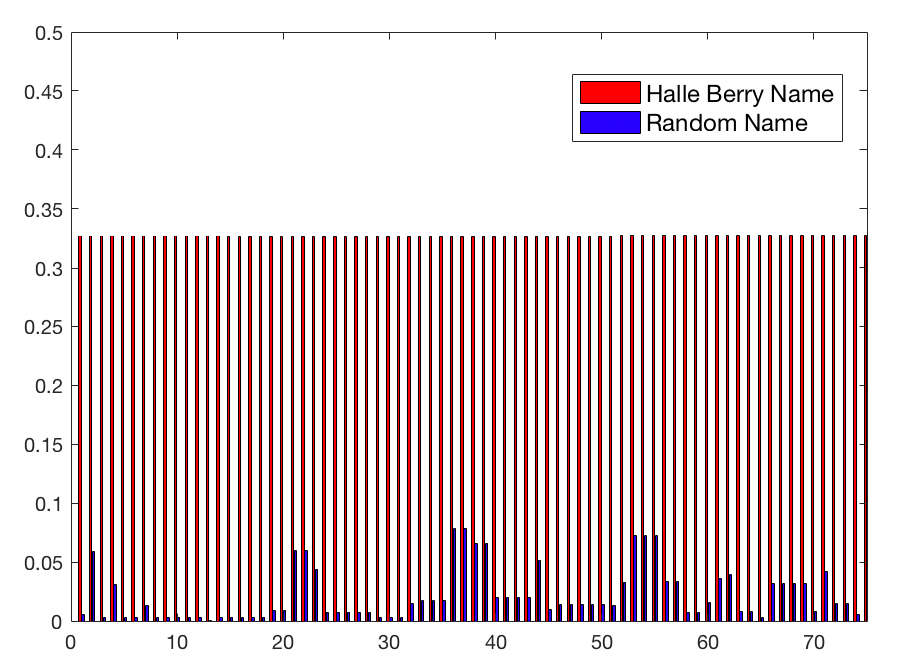}}
\caption{The activation of neuron 326 is shown on the test set when using a single modality as input, (a) Vision (face) only, (b) Text (name) only.  In nearly all test cases, the Halle Berry input activates neuron 326 stronger than a non-Halle Berry input. }
	\label{fig:halleberryneuron2}
\end{figure}

\subsection{Multimodal Representation}
When analyzing the joint representations created by our model, we were fascinated that invariant neurons naturally emerged.  To explore this further, we selected 75 image/text pairs of Halle Berry and 75 random inputs  from our test set and fed them into the network.  We extracted and plotted the multimodal representation at the joint layer of the Deep Sparse Coding model (DSC) and Feed Forward Autoencoder (FFA).  The average activation of these test images can be seen in Figure \ref{fig:halleberryneuron}.  

In the DSC model, we found that neuron 326 actively fires for Halle Berry input but not on other people or other names.  Upon further inspection, using only inputs from one modality (just the face or just the text), N-326 activates with either modality, see Figure \ref{fig:halleberryneuron2}.  Our results demonstrate the invariance of this neuron to modality, an exciting property that naturally emerged from the introduction of biological concepts in the network.  We call N-326 the Halle Berry neuron of our Deep Sparse Coding model.  

In contrast, the top three spikes in the FFA are not multimodal neurons.  Neurons 146, 230, and 275 are the top three activations in the FFA, none of which activate on the text modality.  It appears that the FFA is doing a nonlinear variant of principle component analysis where the model is encoding variance components shared across many image and text inputs.  The FFA is also learning to separate the joint embedding layer as only 27 of the 512 (5.27\%) embedding layer have multimodal representations and respond to both face and name inputs.  This is in stark contrast to our DSC model where 306 of the 512 (59.76\%) neurons activate for both visual and text input.  

The multimodal nature of the DSC neurons has interesting implications.  For example, one could simply threshold the activation of N-326 and get nearly 90\% accuracy on the task of Halle Berry face classification in our test set.  The features extracted are also more informative as we discuss in the next section.

\begin{figure}[th]
	\centering
	{\includegraphics[width=2.01cm]{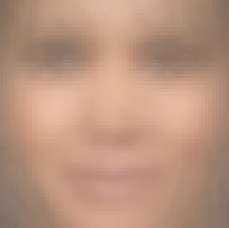}}
	{\includegraphics[width=1.99cm]{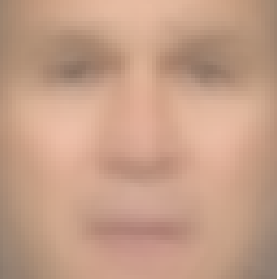}}
	{\includegraphics[width=2cm]{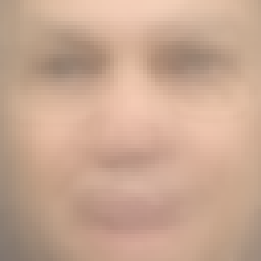}}
	{\includegraphics[width=2cm]{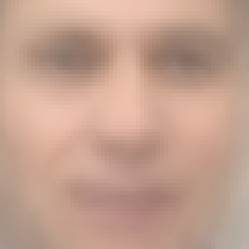}}
	\subfigure[N-326]{\includegraphics[width=2cm]{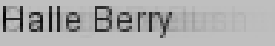}}
	\subfigure[N-493]{\includegraphics[width=2cm]{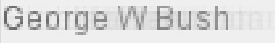}}
	\subfigure[N-121]{\includegraphics[width=2cm]{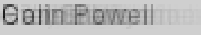}}
	\subfigure[N-220]{\includegraphics[width=2cm]{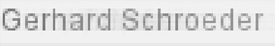}}
\caption{Activity triggered averages corresponding to various neurons in the DSC model.  Neuron 326 is the invariant Halle Berry neuron, which activates strongly to both the face of Halle Berry and the text, ``Halle Berry''.  Other invariant neurons emerged including the George Bush neuron (N-493), Colin Powell neuron (N-121), and Gerhard Schroeder neuron (N-220).   }
	\label{fig:halleberryneuron3}
\end{figure}

Finally, to visualize the activation of the joint embedding neurons i.e. see what the neuron responds to, we use a technique called activity triggered averaging.  We create a compilation of the activation of a neuron at the joint layer multiplied by the input.  We continually add to the compilation and visualize the average input after an epoch of training.  The activity triggered averages of vision and text of selected neurons can be seen in Figure \ref{fig:halleberryneuron3}.

\subsection{Feature Extraction}
\begin{figure}[th]
	\centering
	\subfigure[\textbf{DSC} t-SNE]{\includegraphics[width=4.0cm]{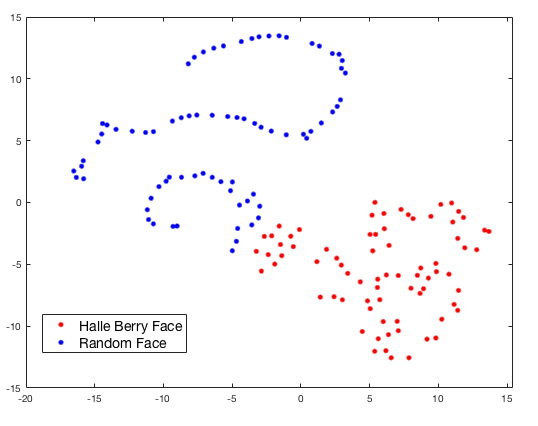}}\hfill
	\subfigure[\textbf{FFA} t-SNE]{\includegraphics[width=4.15cm]{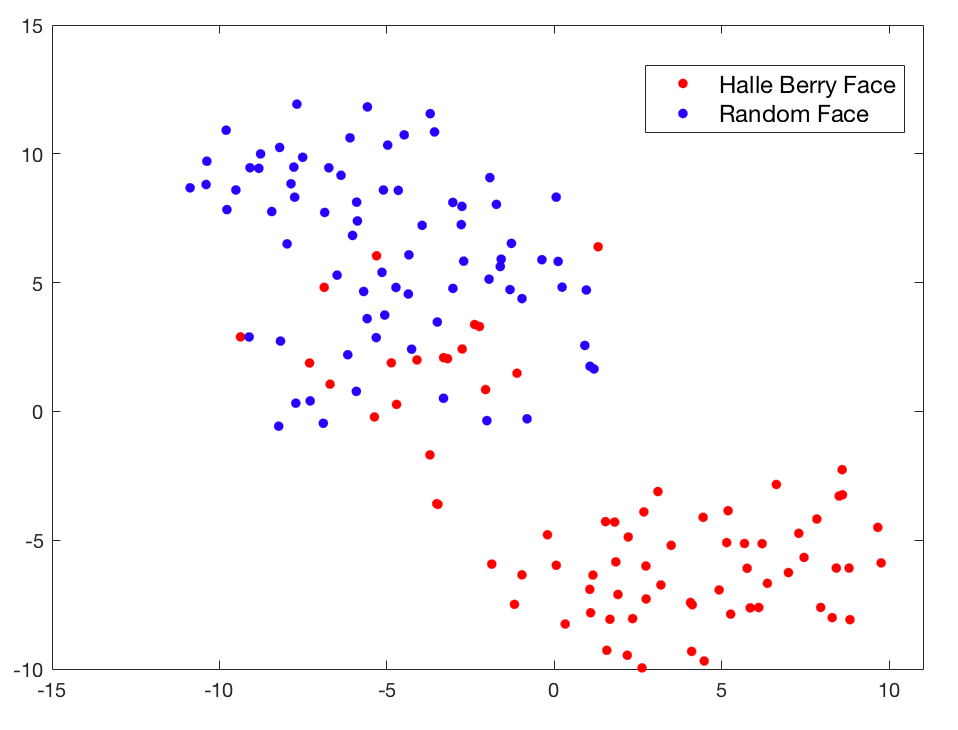}}
\caption{t-SNE plot of the joint embedding layer extracted from our deep sparse coding network (DSC) vs a standard feed-forward autoencoder (FFA).  Our DSC is perfectly separable whereas the FFA has intermingling of classes.  The same parameters (PCA=30, perplexity=30) were used to generate the plots. }
	\label{fig:tsne1}
\end{figure}

Given that our joint embedding produces multimodal neurons, we wanted to explore the uses of the embedding as a feature representation of the data.  The features extracted can be used in classification, clustering, and various other machine learning tasks.  As noted by Glorot et al. \cite{glorot2011deep}, evidence suggests that sparse representations are more linearly separable.  We measure the sparsity of our joint representation in the DSC with the sparsity shown in the FFA.  Our model is on average 20.2\% sparse at the joint layer compared to 47.8\% sparse in the FFA.  Tweaking the FFA's L1 activity regularization to encourage more sparsity resulted in drastically reduced reconstruction performance.  When visualizing the extracted features from the test input, we are able to confirm that the DSC model produces more easily separable features than the FFA, see Figure \ref{fig:tsne1}.

\subsection{Generate Missing Modalities}

\begin{figure}[th]
	\centering
	\includegraphics[width=8cm]{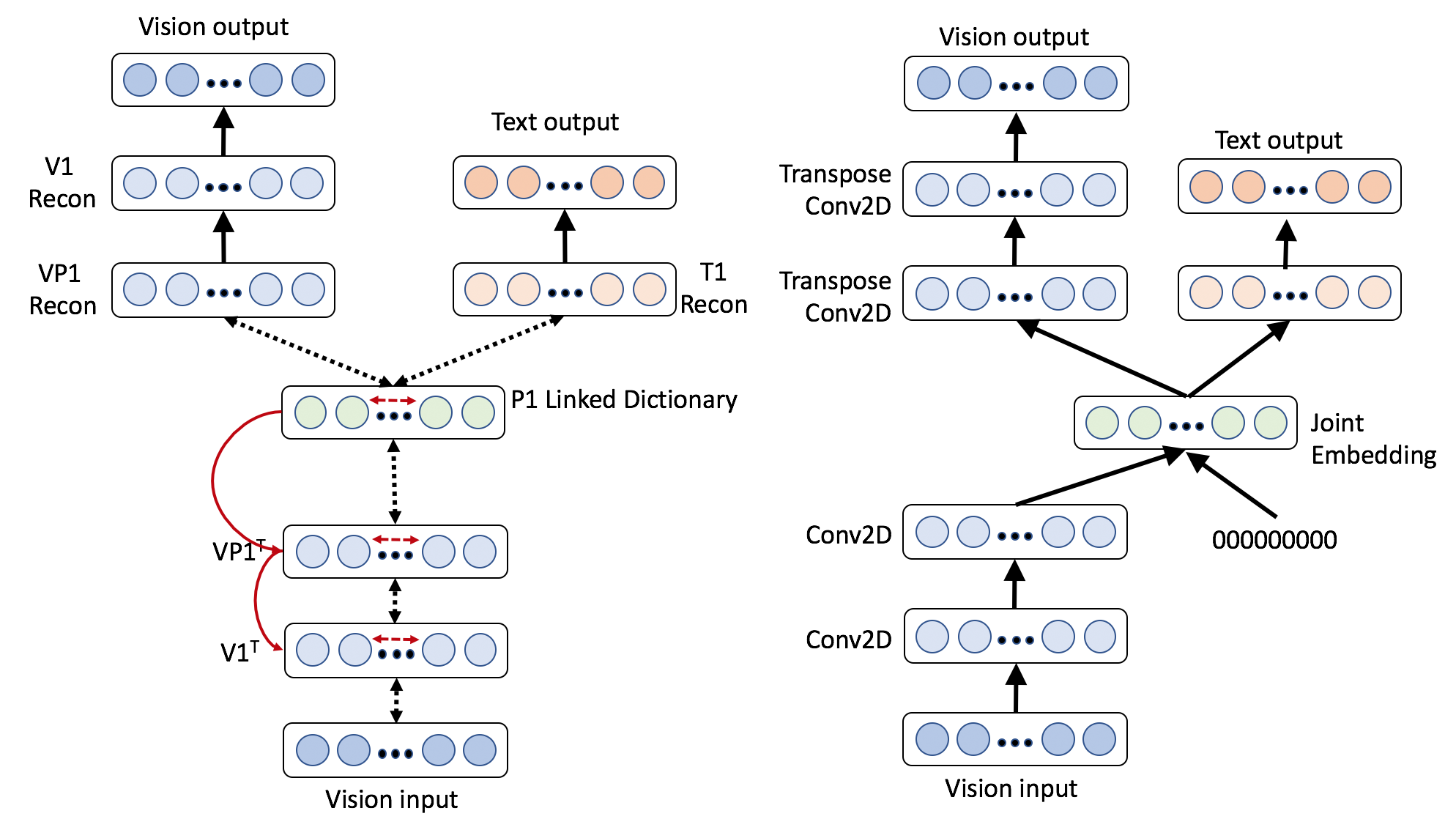}
\caption{Neural network used to generate the text missing modality.  For our DSC model, we can simply remove the input chain, whereas in the FFA model, we fill in the missing input with the zero vector.}
	\label{fig:modelmiss}
\end{figure}

\begin{figure}[th]
	\centering
	\includegraphics[width=7cm]{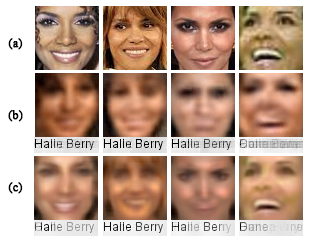}
\caption{Output from the generating network.  (a) The original input image, (b) the face reconstruction and text hallucination from FFA, (c) the face reconstruction and text hallucination from DSC.}
	\label{fig:hallucination}
\end{figure}

In a multimodal network, a common task is the omission of one input to a trained network and forcing it to reconstruct that input as well as generate the missing modality.  Sparse coding as a generative model is naturally adept in ``hallucinating'' a missing modality and we can force the decoder of the feed-forward model to render some output as well.  Figure \ref{fig:modelmiss} shows the removal of the text portion of the input while maintaining the vision and text output.  Since the joint layer in the FFA is an addition of the two streams, we inject the zero vector to simulate a missing input.  Qualitative results can be seen in Figure \ref{fig:hallucination}.  Both the DSC and FFA are able to generate the same number of legible text outputs with relatively high accuracy, approximately 70\%.  Both models fail on the same input images; one example of a failure can be seen in last column of Figure \ref{fig:hallucination}.  However, one notable difference is the in the reconstruction quality of the visual output.  The DSC model reconstructs with greater detail and resemblance to the input.  This a powerful testimony to sparse coding which only trained over the data for 3 epochs versus 25 epochs for the feed-forward autoencoder.

\subsection{Limitations}
Our method is a novel approach to incorporating biologically inspired concepts into a deep neural network and has not been tested on many datasets; however, we believe these concepts make important contributions to the field.   Time and computation in the sparse coding model is the most obvious limitation.  The FFA model is fast and highly optimized, whereas, sparse coding has an expensive inference step.  Although this is a current limitation, specialized hardware is being developed that can do sparse coding with neuromorphic chips on many core meshes \cite{davies2018loihi} and  memristor networks \cite{sheridan2017sparse}.  

\section{Conclusion and Future Work}
In conclusion, we augmented a standard feed-forward autoencoder with biologically inspired concepts of sparsity, lateral inhibition, and top-down feedback.  Inference is achieved by optimization via the LCA sparse solver, rather than an autoencoder model.  Our deep sparse coding model was trained on multimodal data and the joint representation was extracted for comparison against the standard feed-forward encoding.  The joint embedding using our sparse coding model was shown to be more easily separable and robust for classification tasks.  The neurons in our model also had the property of being invariant to modality, with neurons showing activations for both modalities, whereas, the standard feed-forward model simply segregates the modality streams.   

\begin{figure}[h]
	\centering
	\includegraphics[width=8cm]{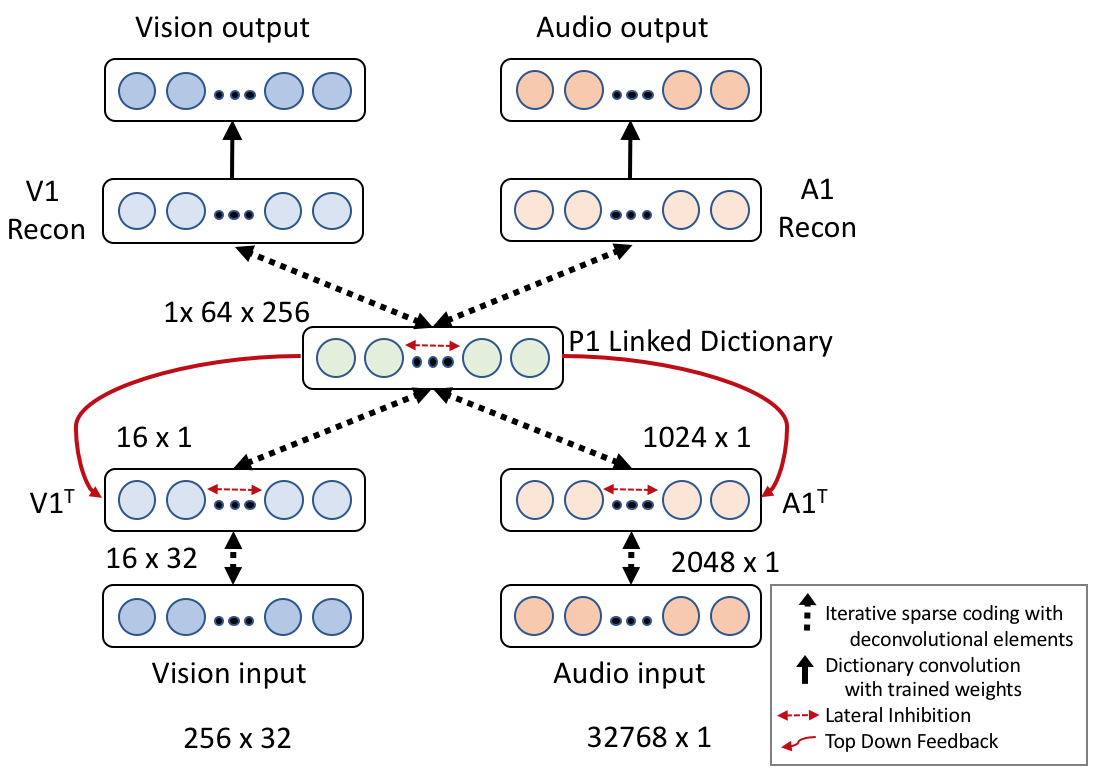}
\caption{Deep Sparse Coding model with raw text image input and raw audio input.  The text is represented by a 256x32 pixel grayscale image while the audio is represented by a 4 second, 32,768 x 1 audio stream sampled at 8 kHz.  Invariant representations of audio and text appear for several concepts.}
	\label{fig:audio}
\end{figure}

Our experimentation illustrates our model using pixels as the input signal on both modality streams; however generally speaking, these can be considered distinct signals with non-overlapping dictionary representations.  As we continue our work, we are experimenting with pixels and audio waveforms in our multimodal deep sparse coding model, see Figure \ref{fig:audio}.   Our model is able to learn neurons that are invariant to audio and text which we plan to explore in future research.  

\begin{figure}[th]
	\centering
	\includegraphics[width=5.8cm]{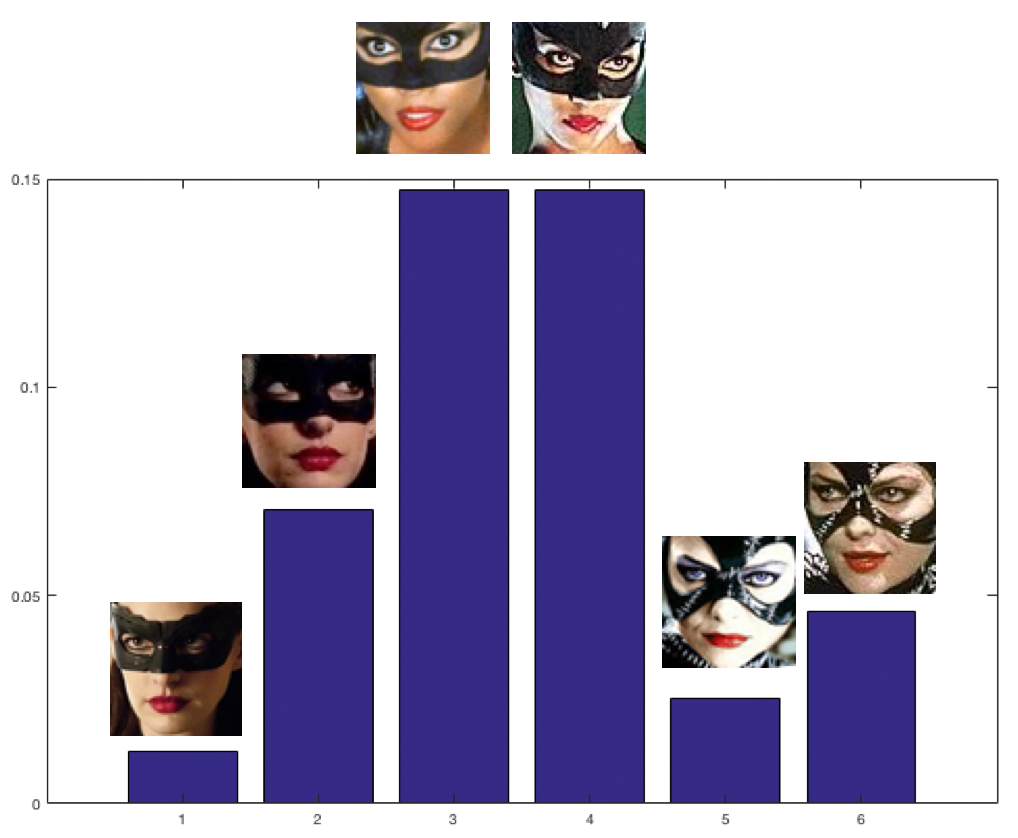}
\caption{Activation on neuron 326 with various pictures of catwoman.  N-326 activates more when tested with Halle Berry as Catwoman versus Anne Hathaway and Michele Pfeiffer.}
	\label{fig:catwoman}
\end{figure}

Finally, for completeness, Quiroga et al. \cite{quiroga2005invariant} noted that the Halle Berry neuron, ``...was also activated by several pictures of Halle Berry dressed as Catwoman, her character in a recent film, but not by other images of Catwoman that were not her''.  As shown in Figure \ref{fig:catwoman}, our model similarly distinguishes between catwomen.

{\small
\bibliographystyle{ieee}
\bibliography{egbib}
}

\end{document}